\documentclass[letterpaper]{article} 
\usepackage[preprint]{aaai2027}  
\usepackage[hyphens]{url}  
\usepackage{graphicx} 
\usepackage{natbib}  
\usepackage{microtype}  
\usepackage{caption}
\usepackage{xr}
\usepackage{algorithm}
\usepackage{algorithmic}
\usepackage{colortbl}
\usepackage{amsmath}
\usepackage{amssymb}
\usepackage{mathtools}
\usepackage{amsthm}

\usepackage{booktabs}
\usepackage{multirow}

\usepackage{xcolor}
\usepackage{pifont}
\usepackage{fontawesome}

\newcommand{\badgelink}[2]{%
  \leavevmode\pdfstartlink attr{/Border [0 0 0]}%
  user{/Subtype /Link /A << /S /URI /URI (#1) >>}%
  #2\pdfendlink%
}
\definecolor{resourcelinkblue}{HTML}{001A80}
\newcommand{\resourcelink}[3]{%
  \badgelink{#1}{%
    \mbox{{\color{black}\fontsize{10.5}{12}\selectfont\mdseries #2}%
    \hspace{0.3em}%
    {\color{resourcelinkblue}\fontsize{10}{12}\selectfont\rmfamily\bfseries #3}}}%
}

\newcommand{\ms}[2]{$#1\pm#2$}
\newcommand{\msb}[2]{\cellcolor[gray]{0.93}$\mathbf{#1\pm#2}$}
\newcommand{\md}[1]{$#1$}

\theoremstyle{plain}
\newtheorem{theorem}{Theorem}

\theoremstyle{definition}

\title{Tools Are Not Islands: Set-Level Tool Retrieval for LLM Agents via Query-Conditioned Hyperedge Prediction}

\author{
    Xinyi Hong\textsuperscript{\rm 1},
    Pinjun Dong\textsuperscript{\rm 2},
    Xinyang Yu\textsuperscript{\rm 2},
    Binyan Jiang\textsuperscript{\rm 2}\corresponding
}

\affiliations{
    \textsuperscript{\rm 1}School of Mathematical Sciences,
    Shanghai Jiao Tong University\\
    \textsuperscript{\rm 2}Department of Data Science and Artificial Intelligence,
    The Hong Kong Polytechnic University\\
    xinyi.hong@sjtu.edu.cn,
    pin-jun.dong@connect.polyu.hk,
    X.Yu59@lse.ac.uk,
    by.jiang@polyu.edu.hk
}

\makeatletter
\g@addto@macro{\@maketitle}{%
  \vspace{-25pt}%
  \begin{center}
  \resourcelink{https://stormwther18-hyset-demo.hf.space}{%
    \raisebox{-0.18\height}{\includegraphics[height=1.15em]{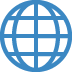}}}{Live Demo}%
  \hspace{2.4em}%
  \resourcelink{https://github.com/stormwther18/HYSET}{\faicon{github}}{GitHub}%
  \hspace{2.4em}%
  \resourcelink{https://huggingface.co/stormwther18}{%
    \raisebox{-0.18\height}{\includegraphics[height=1.25em]{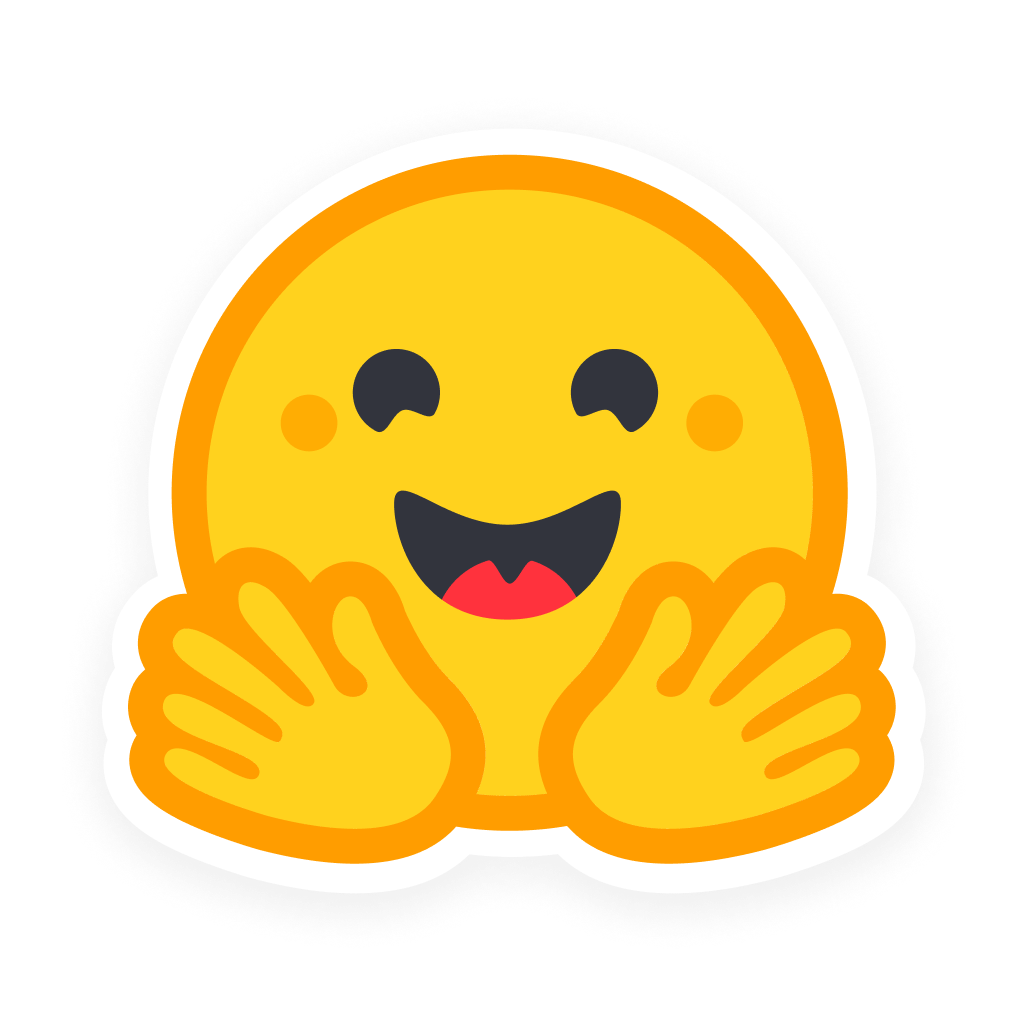}}}{Hugging Face}%
  \end{center}
  \vspace{10pt}%
}
\makeatother

\begin{document}
\maketitle

\begin{abstract}
Large language model (LLM) agents increasingly rely on invoking external tools to complete real-world tasks. Tool retrieval, which selects a small task-relevant subset from a library of thousands of tools before the agent acts, has therefore become a critical component of LLM agent pipelines. However, existing retrievers either score each tool in isolation or assemble the tool set sequentially, so the joint utility of a candidate set is never evaluated as a whole. In this paper, we propose \textbf{HYSET}, short for \textbf{HY}peredge-based \textbf{SE}t-level \textbf{T}ool retrieval. Our contributions are threefold: (i) we formulate tool retrieval as query-conditioned hyperedge prediction on a tool co-invocation hypergraph, under which the tool set itself becomes the unit of scoring and most existing retrieval paradigms reduce to restricted instances; (ii) we capture size-dependent tool compatibility through cardinality-specific interactions; and (iii) we design HYSET as a pre-selection module requiring no modification to the downstream agent. Experiments on ToolBench demonstrate that HYSET consistently outperforms state-of-the-art baselines in both tool retrieval performance and end-to-end task success. Beyond the in-domain setting, HYSET further supports zero-shot/few-shot transfer, generalizing to held-out tools/categories and unseen domains with minimal supervision.
\end{abstract}

\section{Introduction}
\label{sec:intro}

Large language models (LLMs) are increasingly deployed as autonomous agents capable of invoking external tools to complete real-world tasks~\citep{schick2023toolformer,shen2023hugginggpt,patil2024gorilla,yao2023react}. As agents grow more capable and their invocation demands expand, tool libraries have scaled from dozens of handcrafted functions to large-scale API ecosystems containing thousands of real-world endpoints. ToolBench~\citep{qin2024toolllm}, for instance, includes 16{,}464 API endpoints spanning 49 categories from the RapidAPI Hub, a scale that renders exhaustive in-context presentation of every tool description impractical. One might expect that increasingly long context windows would eventually eliminate the need for explicit tool selection. However, empirical evidence shows that the effective use of information degrades sharply when context exceeds tens of thousands of tokens~\citep{liu2024lost}, and injecting the full tool library into every prompt would incur prohibitive latency and cost at agent scale. Moreover, the difficulty is not one of scale alone. Real-world tasks rarely depend on a single API, and a query is typically resolved by several APIs invoked jointly~\citep{qin2024toolllm,qu2024towards}, so what must be retrieved is not a ranked list of individually relevant tools but a jointly useful tool set. Tool retrieval, which selects such a small task-relevant subset from a large API library before the agent acts, therefore remains a persistent infrastructure bottleneck rather than a transitional one.

\begin{figure*}[t]
    \centering
    \includegraphics[
        width=1\textwidth,
        trim=1cm 6cm 0 5cm,
        clip
    ]{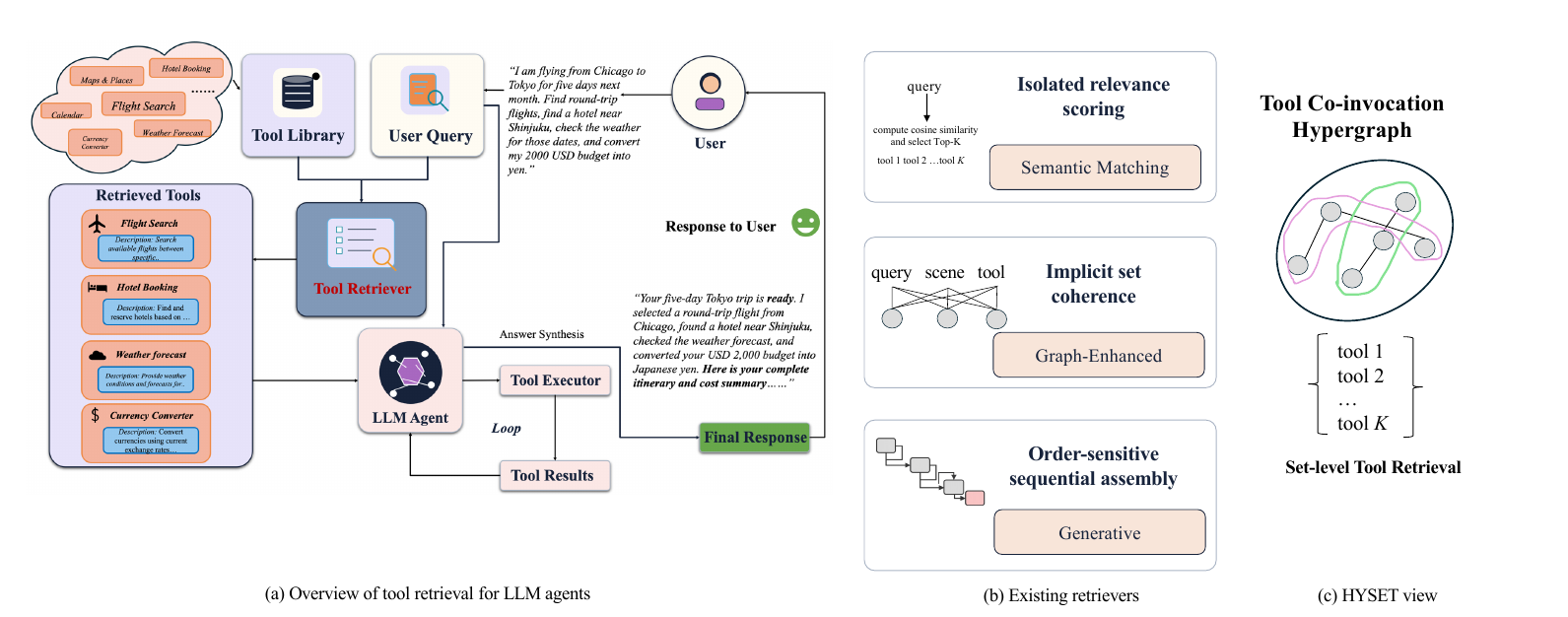}
\caption{\textbf{Overview and motivation of HYSET.}
(a) Tool-retrieval pipeline; (b) limitations of existing retrievers;
and (c) our set-level view as query-conditioned hyperedge prediction.}
    \label{fig:1}
\end{figure*}

Existing tool retrievers address this bottleneck through three representative paradigms. The most common paradigm, semantic-matching retrieval, scores each API independently against the query. Sparse methods such as BM25~\citep{robertson2009probabilistic} rank APIs by lexical overlap, while dense bi-encoders such as Contriever~\citep{izacard2021unsupervised} and Sentence-BERT-style models~\citep{reimers2019sbert,qin2024toolllm} score APIs by cosine similarity in a shared embedding space. All such methods treat each tool as an independently scorable item, so the score of a candidate set is simply an aggregate of per-tool signals. This treatment breaks down whenever tool utility is a joint property of the set rather than a sum of individual values. A tool that ranks low in isolation may become indispensable alongside other selected APIs, and a group of individually high-scoring tools may collectively fail to cover the full scope of a task. This gap is evident in practice. On ToolBench, a fine-tuned Contriever retriever reported by \citet{qu2024towards} attains a Recall@3 of 68.6\%, yet its COMP@3, the fraction of queries whose complete required tool set is covered by the top-3 results, is only 39.7\%. To move beyond such independent scoring, the other two paradigms have recently been explored. Graph-enhanced retrievers such as COLT~\citep{qu2024towards} augment semantic matching with dual-view collaborative learning over a bipartite query-scene-tool graph, improving the recovery of the complete required tool set. Nevertheless, their collaborative signal is distilled into per-tool embeddings during training, and inference still reduces to independent top-$k$ ranking, so set-level coherence is only approximated rather than scored explicitly. Generative retrievers such as ToolGen~\citep{wang2024toolgen}, building on generative retrieval~\citep{tay2022transformer}, instead emit tool identifiers directly from the language model. However, the tool set is assembled step by step from local conditional probabilities under a sequence-likelihood objective, so complete candidate sets are never explicitly compared, and whether the assembled set is coherent and complete can only be judged after generation ends. Figure~\ref{fig:1} contrasts these paradigms with our set-level view. Supplementary Material H gives extended related work.

Across all three paradigms, tools are either scored independently or generated sequentially, and no candidate set is ever assessed as a whole. We therefore argue that \textbf{tools are not islands, and tool retrieval is inherently a set-level problem}. This position raises two questions. \emph{\textbf{(Q1)} Can tool sets be scored as a whole, and does this improve retrieval? \textbf{(Q2)} Do tool co-invocation patterns vary with set size, and does modeling these differences help?} Both are grounded in a representative ToolBench example. For the travel-planning query ``\textit{I am flying from Chicago to Tokyo for five days next month. Find round-trip flights, book a hotel near Shinjuku, check the weather for those dates, and convert my 2{,}000 USD budget into yen}'', whose ground-truth tools are $\{\textit{Flight},\textit{Hotel},\textit{Weather},\textit{Currency}\}$, flight terms dominate the wording and a fine-tuned dense retriever scores \textit{Flight}, \textit{CheapFlight}, \textit{FlightTracker}, \textit{Hotel}, \textit{Weather} and \textit{Currency} at $0.92$, $0.89$, $0.87$, $0.80$, $0.55$ and $0.32$. Top-4 retrieval therefore returns $\{\textit{Flight},\textit{CheapFlight},\textit{FlightTracker},\textit{Hotel}\}$ and leaves two subtasks uncovered. The failure is structural rather than a matter of calibration. Once \textit{Flight} is selected, a second flight API contributes almost nothing, yet each per-tool score is assigned in ignorance of what else has been selected, which is the concern of \emph{\textbf{(Q1)}} and is visible only when tools are scored as a whole. However, scoring the set jointly is still not enough, because the interaction among tools must itself depend on the cardinality of the set. A currency converter and a weather lookup belong together in the four-tool request above, but in a two-tool task they are co-invoked only under a contrived query such as ``\textit{convert my budget into yen and tell me whether it will rain in Tokyo}''. On ToolBench the two co-occur in $0.4\%$ of two-tool sets but in $23\%$ of four-tool sets, a gap far wider than the mechanical increase in pair count with set size would explain. This underlies \emph{\textbf{(Q2)}} and motivates cardinality-specific interaction modeling.

We therefore propose \textbf{HYSET}, short for \textbf{HY}peredge-based \textbf{SE}t-level \textbf{T}ool retrieval. Our main contributions are summarized as follows:
\begin{itemize}
\item \textbf{Formulation.} To the best of our knowledge, we are the first to recast tool retrieval for LLM agents as query-conditioned hyperedge prediction over a tool co-invocation hypergraph, making the tool set itself the unit of scoring. We further provide a unified view where existing tool-retrieval paradigms arise as restricted instances.

\item \textbf{Method.} We design HYSET as a pre-selection module requiring no modification to the downstream agent, which scores candidate tool sets via cardinality-specific interaction matrices so that tool compatibility can vary with set size.

\item \textbf{Experiments.} On ToolBench, HYSET consistently outperforms strong baselines from all three paradigms, with relative improvements of \textbf{11.6\%} in COMP@5 and up to \textbf{13.1\%} in end-to-end pass rate over the strongest baselines. It further transfers zero-shot to held-out tools and categories, and recovers \textbf{93.2\%} of fully supervised performance with only \textbf{5} labeled examples per category.
\end{itemize}

\section{Methodology}
\label{sec:metho}

\begin{figure*}[t]
    \centering
    \includegraphics[
        width=1\textwidth,
        trim=1.5cm 7.8cm 0.6cm 3.6cm,
        clip
    ]{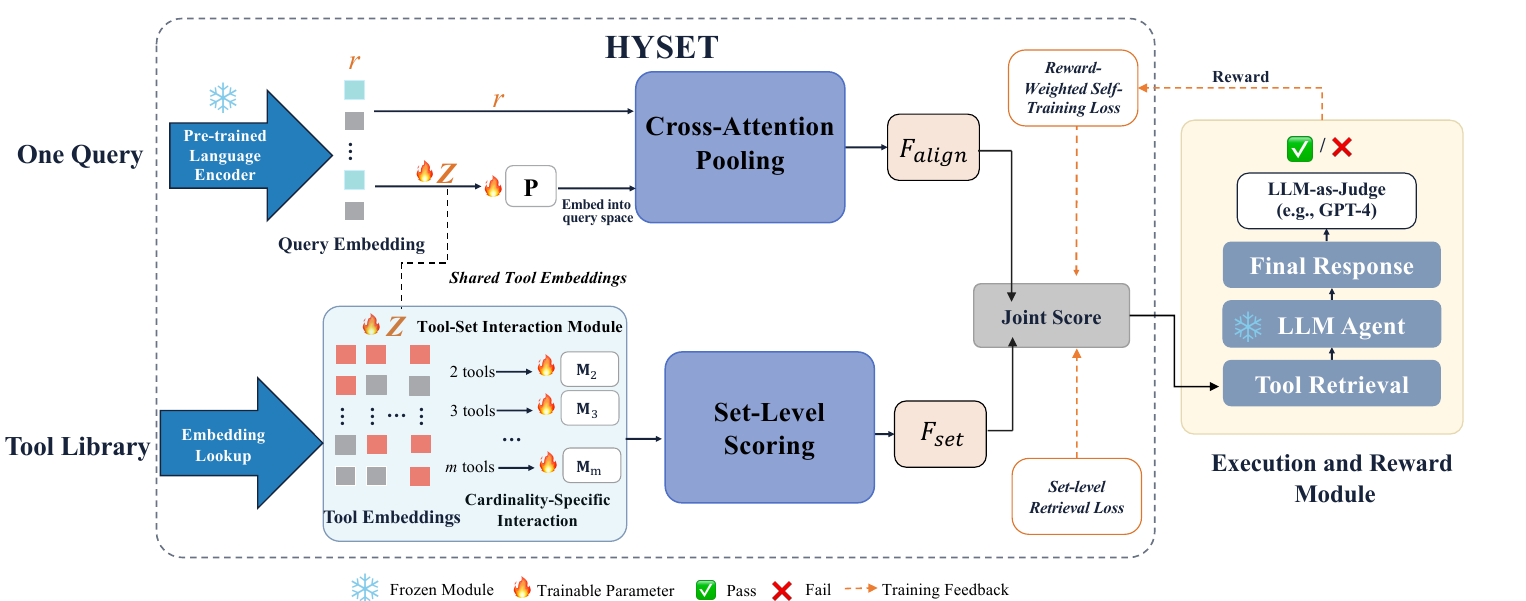}
\caption{\textbf{HYSET framework.} HYSET consists of two components: query-set alignment, measuring relevance to the input query, and set-level scoring, treating each candidate set as a hyperedge and modeling cardinality-specific interactions among its tools.}
    \label{fig:2}
\end{figure*}

\subsection{Preliminaries}
\label{sec:prelim}
Let $\mathcal{T}=\{t_{1},\ldots,t_{N_{\mathcal{T}}}\}$ be a tool library. We assume each query $x$ in the natural-language query space $\mathcal{X}$ can be fulfilled by jointly invoking a subset $E\subseteq\mathcal{T}$, and we assume access to a training set $\mathcal{D}_{\mathrm{tr}}=\{(x_i,E_i^\star)\}_{i=1}^{N}$, where $E_i^\star$ is the annotated tool set for $x_i$. Each $E_i^\star$ is the ground truth for $x_i$ but need not be the unique feasible set, and it is unordered because invocation order is decided by the downstream agent rather than by the retriever. These sets admit a hypergraph representation~\citep{battiston2020networks} that we call the tool co-invocation hypergraph $\mathcal{H}=(\mathcal{V},\mathcal{E})$, whose node set $\mathcal{V}=\mathcal{T}$ collects all tools and whose hyperedges $\mathcal{E}$ are the observed $E_i^\star$, so the unit of supervision shifts from the relevance of an individual tool to the joint utility of an entire set. Writing $M=\max_{i}|E_i^\star|$ for the largest observed tool-set size, the admissible candidate hyperedge space $\mathcal{E}_M$ collects all $E\subseteq\mathcal{V}$ with $1\le|E|\le M$, assuming that tool sets required at inference do not exceed $M$.

\subsection{Problem Formulation}
\label{sec:problemfor}

Set-level tool retrieval seeks a mapping $\mathcal{R}:\mathcal{X}\rightarrow\mathcal{E}_M$ returning the complete tool set for a query. Since \(\mathcal{V}=\mathcal{T}\), every candidate set \(E\in\mathcal{E}_M\) is a hyperedge over \(\mathcal{V}\), so the task is query-conditioned hyperedge prediction. We model it through a parametric scoring family $F_\theta:\mathcal{X}\times\mathcal{E}_M\rightarrow\mathbb{R}$, where \(F_\theta(x,E)\) measures the joint utility of selecting \(E\) for \(x\), and estimate $\widehat{\theta}=\arg\min_{\theta\in\Theta}\mathcal{L}(\theta;\mathcal{D}_{\mathrm{tr}})$ with \(\mathcal{L}\) the empirical loss of Section~\ref{sec:hots}. For a new query \(x_{\mathrm{new}}\), the learned scoring function $\widehat{F}=F_{\widehat{\theta}}$ predicts
\begin{equation}
\widehat{E}(x_{\mathrm{new}})
=
\arg\max_{E\in\mathcal{E}_M}
F_{\widehat{\theta}}(x_{\mathrm{new}},E).
\label{eq:pre}
\end{equation}
The resulting set $\widehat{E}(x_{\mathrm{new}})$ is subsequently provided to the frozen downstream LLM agent to complete the user query.

\subsection{A Unified View of Tool Retrieval}
\label{sec:unified}

Under the hypergraph formulation and scoring family introduced in
Section~\ref{sec:problemfor}, major existing tool-retrieval paradigms can be subsumed by a unified set-scoring framework.
 \begin{itemize}
\setlength{\itemsep}{0pt}\setlength{\parsep}{0pt}\setlength{\topsep}{2pt}\setlength{\partopsep}{0pt}
\item[(i)] \textbf{Semantic-matching retrievers} induce
\(F_{\mathrm{match}}:\mathcal{X}\times\mathcal{E}_M\rightarrow\mathbb{R}\)
with $F_{\mathrm{match}}(x,E)
=
\sum_{t\in E}
F_{\mathrm{match}}(x,\{t\})
=
\sum_{t\in E}
s_{\mathrm{match}}(x,t)$,
where \(\{t\}\in\mathcal{E}_M\) is the singleton hyperedge associated with tool \(t\), and
\(s_{\mathrm{match}}(x,t)\)
is typically a lexical or dense query-tool matching score.

\item[(ii)] \textbf{Graph-enhanced retrievers} induce
\(F_{\mathrm{graph}}:\mathcal{X}\times\mathcal{E}_M\rightarrow\mathbb{R}\)
with $F_{\mathrm{graph}}(x,E)
=
\sum_{t\in E}
F_{\mathrm{graph}}(x,\{t\})
=
\sum_{t\in E}
s_{\mathcal{G}}(x,t)$,
where
\(s_{\mathcal{G}}(x,t)\)
is typically the similarity between a query embedding and a graph-contextualized embedding of the individual tool \(t\).

\item[(iii)] \textbf{Generative retrievers} induce
\(F_{\mathrm{gen}}:\mathcal{X}\times\mathcal{E}_M\rightarrow\mathbb{R}\)
with $F_{\mathrm{gen}}(x,E)
=
\log
\sum_{\pi\in\mathfrak{S}(E)}
p_\phi(\pi,\mathrm{EOS}\mid x)$,
where \(\mathfrak{S}(E)\) collects the \(|E|!\) orderings of \(E\), and
\(p_\phi\)
is typically the autoregressive probability of a language model over tool-identifier tokens. Its factorization is given in Supplementary Material A.2.
\end{itemize}
The formulations above provide a unified mathematical view of existing paradigms. Supplementary Material A.2 derives the three induced scores and A.3 shows that they form a strict hierarchy of interaction orders.

\begin{algorithm}[t]
\caption{Training and Inference of HYSET}
\label{alg:1}
\textbf{Training input}: Tool node set \(\mathcal{V}\), training set \(\mathcal{D}_{\mathrm{tr}}=\{(x_i,E_i^\star)\}_{i=1}^{N}\), frozen query encoder \(\mathbf{r}(\cdot)\), frozen agent \(\mathcal{A}\), candidate-pool size \(K_{\mathrm{neg}}\), maximum cardinality \(M\), execution limit \(R\), and weights \(\eta,\lambda\)\\
\textbf{Inference input}: Query \(x_{\mathrm{new}}\) and shortlist sizes \(K_1<K_{\mathrm{pool}}\)\\
\textbf{Output}: Learned scoring function \(\widehat{F}\) and predicted tool set \(\widehat{E}(x_{\mathrm{new}})\)
\begin{algorithmic}[1]
\item[] \textbf{Training phase}
\REPEAT
    \STATE Sample a minibatch index set \(\mathcal{B}\subseteq\{1,\ldots,N\}\)
    \FOR{each \(i\in\mathcal{B}\)}
        \STATE Construct \(\mathcal{C}_i\subseteq\mathcal{E}_M\) of size \(K_{\mathrm{neg}}\) using the negative-sampling procedure in Section~\ref{sec:hots}
        \STATE Compute \(\widehat{E}_i\) from \(\mathcal{C}_i\) via Eqs.~\eqref{eq:F} and~\eqref{eq:topscore}
        \STATE Run \(\mathcal{A}\) on \(x_i\) using DFSDT~\citep{qin2024toolllm} with tool set \(\widehat{E}_i\) for at most \(R\) steps
        \STATE Obtain the task reward \(\rho_i\) from the execution result
    \ENDFOR
    \STATE Update \(\theta\) via Eq.~\eqref{eq:constrained-training}
\UNTIL{convergence}
\STATE Obtain \(\widehat{\theta}\leftarrow\theta\) and \(\widehat{F}=F_{\widehat{\theta}}\)
\item[] \textbf{Inference phase}
\STATE Construct the \(K_{\mathrm{pool}}\)-tool shortlist \(\mathcal{S}\) using Eqs.~\eqref{eq:singleton} and~\eqref{eq:expand}
\STATE Compute \(\widehat{E}(x_{\mathrm{new}})\) via Eq.~\eqref{eq:inference-rerank}
\STATE \textbf{return} \(\widehat{F}\) and \(\widehat{E}(x_{\mathrm{new}})\)
\end{algorithmic}
\end{algorithm}
\subsection{The Framework of HYSET}
\label{sec:hots}
\paragraph{Scoring Function Parameterization.}
We parameterize the scoring function \(F_\theta:\mathcal{X}\times\mathcal{E}_M\to\mathbb{R}\) introduced in Section~\ref{sec:problemfor} through the decomposition
\begin{equation}
F_\theta(x_i,E)
=
F_{\mathrm{set}}(E)
+
F_{\mathrm{align}}(x_i,E),
\label{eq:F}
\end{equation}
where \(F_{\mathrm{set}}\) models the internal interaction among the tools in \(E\) independently of the query, and \(F_{\mathrm{align}}\) models the alignment between \(x_i\) and \(E\).
First, motivated by latent-space models of hypergraph data~\citep{turnbull2024latent,wu2024general,hong2026hyvint}, we parameterize the set-interaction term \(F_{\mathrm{set}}\) as
\begin{equation}
F_{\mathrm{set}}(E)
=
\sum_{1\leq a<b\leq m}
\mathbf{z}_{j_a}^{\top}
\mathbf{M}_m
\mathbf{z}_{j_b},
\label{eq:Hm}
\end{equation}
where \(E=\{t_{j_1},\ldots,t_{j_m}\}\in\mathcal{E}_M\) is a candidate hyperedge of cardinality \(m=|E|\), \(\mathbf{z}_{j_a}\in\mathbb{R}^{d_z}\) is the learnable embedding of tool node \(t_{j_a}\), \(\mathbf{Z}\in\mathbb{R}^{|\mathcal{V}|\times d_z}\) collects all tool embeddings as rows, and \(\mathbf{M}_m\in\mathbb{R}^{d_z\times d_z}\) is a symmetric interaction matrix shared across candidate hyperedges of cardinality \(m\). A larger \(F_{\mathrm{set}}(E)\) indicates that the tools of \(E\) better conform to the learned patterns of co-invocation and functional complementarity. For \(m\geq2\), the pair contribution \(\mathbf{z}_{j_a}^{\top}\mathbf{M}_m\mathbf{z}_{j_b}\) is allowed to vary with the cardinality of the set containing it, which is precisely what \(\{\mathbf{M}_m\}_{m=2}^{M}\) is introduced to model.
To formalize the distinction, we introduce the notion of a \textit{pairwise-decomposable} set function based on the M\"obius transform~\cite{grabisch2000equivalent}. Specifically, for a given function $f:\mathcal{E}_M\cup\{\varnothing\}\rightarrow\mathbb{R}$, the M\"obius transform and its inverse are defined as
\begin{equation*}
\tilde{f}(E) := \sum_{T \subseteq E} (-1)^{|E|-|T|} f(T),\quad
f(E) = \sum_{T \subseteq E} \tilde{f}(T).
\end{equation*}
Intuitively, the M\"obius coefficient $\tilde{f}(T)$, whose interaction order is $|T|$, measures the joint contribution attributable specifically to $T$ after subtracting the contributions of its proper subsets. Hence, we call $f$ \emph{pairwise-decomposable} if $\tilde{f}(E)=0$ whenever $3\leq|E|\leq M$, meaning that it contains no interactions above order two. For instance, for any $E\in\mathcal{E}_M$ with $|E|=3$, its third-order interaction is
$
\tilde{F}_{\mathrm{set}}(E)
=
\sum_{\{t_i,t_j\}\subseteq E}
\mathbf{z}_{i}^{\top}(\mathbf{M}_3-\mathbf{M}_2)\mathbf{z}_{j}.
$
Thus, $\mathbf{M}_3-\mathbf{M}_2$ induces the third-order effect without an explicit three-way tensor. More generally, we have the following theorem:
\begin{theorem}
\label{thm:order}
Assume $|\mathcal{V}|\ge M+2$. The set function $F_{\mathrm{set}}:\mathcal{E}_M\cup\{\varnothing\}\rightarrow\mathbb{R}$
is pairwise-decomposable if and only if
\begin{equation*}
\mathbf{z}_{i}^{\top}\mathbf{M}_2\mathbf{z}_{j}
=
\cdots
=
\mathbf{z}_{i}^{\top}\mathbf{M}_M\mathbf{z}_{j},
\qquad \forall\, t_i,t_j\in\mathcal{V},\ i\ne j.
\end{equation*}
\end{theorem}

We prove Theorem~\ref{thm:order} in Supplementary Material A.4. The theorem identifies the exact role of the cardinality-specific matrices. \(F_{\mathrm{set}}\) reduces to a fixed pairwise model precisely when every pair receives the same score at all cardinalities from \(2\) to \(M\). Otherwise, its cardinality dependence induces interactions above order two. Thus, although \(F_{\mathrm{set}}\) is parameterized as a sum over pairs, it can represent structured joint effects at orders up to \(M\) and thereby capture multi-tool compatibility that an ordinary graph with fixed pairwise edge weights cannot express. These structured effects are generated by \(O(Md_z^{2})\) parameters, rather than the \(O(d_z^{M})\) parameters required by an unrestricted explicit \(M\)-way tensor.
Second, we parameterize the query-set alignment \(F_{\mathrm{align}}\) as
\begin{equation}
F_{\mathrm{align}}(x_i,E)
=
\mathbf{r}(x_i)^{\top}\mathbf{s}(x_i,E),
\label{eq:Ftheta2}
\end{equation}
where \(\mathbf{r}:\mathcal{X}\rightarrow\mathbb{R}^{d_r}\) is a frozen pre-trained language model that maps query \(x_i\) to its embedding \(\mathbf{r}(x_i)\)~\citep{reimers2019sbert}, and \(\mathbf{s}:\mathcal{X}\times\mathcal{E}_M\rightarrow\mathbb{R}^{d_r}\) produces a query-conditioned representation \(\mathbf{s}(x_i,E)\) of candidate hyperedge \(E\). Specifically, we compute \(\mathbf{s}(x_i,E)\) using cross-attention pooling with \(\mathbf{r}(x_i)\) as the query and \(\{\mathbf{P}\mathbf{z}_{j_k}\}_{k=1}^{m}\) as the keys and values~\citep{vaswani2017attention},
\begin{equation}
\mathbf{s}(x_i,E)
=
\sum_{k=1}^{m}
\alpha_k(x_i,E)\mathbf{P}\mathbf{z}_{j_k},
\label{eq:attn}
\end{equation}
where $\alpha_k(x_i,E)=
\frac{\exp\!\left(\ell(x_i,t_{j_k})\right)}
{\sum_{q=1}^{m}\exp\!\left(\ell(x_i,t_{j_q})\right)}$ is the normalized attention weight assigned to \(t_{j_k}\) within \(E\), and $\ell(x_i,t_{j_k})=
\mathbf{r}(x_i)^{\top}\mathbf{P}\mathbf{z}_{j_k}$ is the matching score between query \(x_i\) and tool \(t_{j_k}\).
The matrix \(\mathbf{P}\in\mathbb{R}^{d_r\times d_z}\) maps each tool embedding into the query space. Substituting Eq.~\eqref{eq:attn} into Eq.~\eqref{eq:Ftheta2} gives
\begin{equation}
F_{\mathrm{align}}(x_i,E)
=
\sum_{k=1}^{m}
\alpha_k(x_i,E)\ell(x_i,t_{j_k}).
\label{eq:alignment-interpretation}
\end{equation}
For \(m\geq2\), each attention weight depends on all tools in \(E\), so the alignment term is itself a query-conditioned set-level score (Supplementary Material A.5). The trainable parameters are \(\theta=(\mathbf{Z},\{\mathbf{M}_m\}_{m=2}^{M},\mathbf{P})\), with \(\mathbf{Z}\) shared by the two terms.
\paragraph{Training Objective.}
In practice, we construct the training objective of HYSET from negative-sampled set supervision and execution feedback, as depicted in Figure~\ref{fig:2} and summarized in Algorithm~\ref{alg:1}. Since the admissible space of Section~\ref{sec:prelim} contains \(|\mathcal{E}_M|=\sum_{m=1}^{M}\binom{|\mathcal{V}|}{m}\) candidate hyperedges, exhaustive training is infeasible, and we use negative sampling both to build tractable candidate pools and to supply contrasting tool sets for the retrieval loss. For each query \(x_i\) with ground-truth hyperedge \(E_i^\star\), we form the pool $\mathcal{C}_i=\{E_i^\star\}\cup\mathcal{N}_i$ with $\mathcal{N}_i\subseteq\mathcal{E}_M\setminus\{E_i^\star\}$ and $|\mathcal{N}_i|=K_{\mathrm{neg}}-1$ distinct negatives, drawn from three sources in the fixed proportion \(50\%/30\%/20\%\): size-matched sets sampled uniformly from the \(|E_i^\star|\)-tool subsets of \(\mathcal{V}\), in-batch negatives taken from the ground-truth sets of the other queries of the minibatch~\citep{oord2018representation}, and hard negatives obtained by replacing one or two tools of \(E_i^\star\) with their nearest neighbors under \(\mathbf{Z}\)~\citep{karpukhin2020dpr,robinson2020contrastive}. Supplementary Material A.8 states the sampling procedure in full, gives an adaptive mixture recovering this one as the constant special case, and measures the resulting false-negative rate. Supplementary Material C.8 sweeps the proportion. Given \(\mathcal{C}_i\), we score each candidate using Eq.~\eqref{eq:F} and select
\begin{equation}
\widehat{E}_i
=
\arg\max_{E\in\mathcal{C}_i}
F_\theta(x_i,E),
\label{eq:topscore}
\end{equation}
The frozen LLM agent \(\mathcal{A}\) then processes \(x_i\) under the DFSDT search procedure~\citep{qin2024toolllm} with its tool set restricted to \(\widehat{E}_i\) for at most \(R\) steps, and its answer \(\widehat{y}_i\) receives the execution reward $\rho_i=\operatorname{ExecScore}(x_i,\widehat{y}_i)\in[0,1]$, whose definition, configuration and refresh schedule are given in Supplementary Material A.9. The candidate pool induces the retrieval loss
\begin{equation}
\mathcal{L}_{\mathrm{ret}}(\theta)
=
-\sum_{i\in\mathcal{B}}
\log
\frac{\exp F_\theta(x_i,E_i^\star)}
{\sum_{E\in\mathcal{C}_i}\exp F_\theta(x_i,E)},
\label{eq:Lret}
\end{equation}
which raises \(F_\theta(x_i,E_i^\star)\) relative to \(F_\theta(x_i,E)\) for every \(E\in\mathcal{N}_i\). We further use execution feedback to reinforce successful model-selected sets through the reward-weighted self-training loss
\begin{equation}
\mathcal{L}_{\mathrm{self}}(\theta)
=
-\sum_{i\in\mathcal{B}}
\rho_i
\log
\frac{\exp F_\theta(x_i,\widehat{E}_i)}
{\sum_{E\in\mathcal{C}_i}\exp F_\theta(x_i,E)},
\label{eq:Ltask}
\end{equation}
where \(\widehat{E}_i\) and \(\rho_i\) are treated as constants within each parameter update. 
Combining Eqs.~\eqref{eq:Lret} and~\eqref{eq:Ltask}, the full objective is
\begin{equation}
\mathcal{L}(\theta)
=
\mathcal{L}_{\mathrm{ret}}(\theta)
+
\eta\mathcal{L}_{\mathrm{self}}(\theta)
+
\lambda
\sum_{m=2}^{M}
\|\mathbf{M}_m\|_F^2,
\label{eq:L}
\end{equation}
where \(\eta>0\) controls the execution-feedback term and \(\lambda>0\) controls the regularization of the cardinality-specific interaction matrices. We estimate
\begin{equation}
\widehat{\theta}
=
\arg\min_{\theta\in\Theta}
\mathcal{L}(\theta)
\quad
\mathrm{s.t.}
\quad
\|\mathbf{z}_j\|_2=1
\quad
\forall\,t_j\in\mathcal{V},
\label{eq:constrained-training}
\end{equation}
where \(\|\mathbf{z}_j\|_2=1\) fixes the scale of each tool embedding.
\begin{table*}[t]
\centering
\small
\setlength{\tabcolsep}{1mm}
\renewcommand{\arraystretch}{1.15}

\begin{tabular}{c c c c c cc}
\toprule
\multirow{2}{*}{Method}
& \multirow{2}{*}{Sup.}
& \multirow{2}{*}{Recall@5 $\uparrow$}
& \multirow{2}{*}{NDCG@5 $\uparrow$}
& \multirow{2}{*}{COMP@5 $\uparrow$}
& \multicolumn{2}{c}{Pass Rate $\uparrow$} \\
\cmidrule(lr){6-7}
& & & & & GPT-4 & Human \\
\midrule

BM25 & A
& \md{41.02}
& \md{39.68}
& \md{22.38}
& \ms{19.21}{0.44}
& \ms{15.37}{1.47} \\

Contriever & A
& \md{41.08}
& \md{38.30}
& \md{22.04}
& \ms{18.80}{0.41}
& \ms{14.17}{1.42} \\

ToolLLaMA-Ret & A
& \md{68.88}
& \md{70.12}
& \md{61.11}
& \ms{64.05}{0.68}
& \ms{61.83}{1.98} \\

ToolRerank & A
& \ms{80.71}{0.17}
& \ms{81.63}{0.08}
& \ms{69.93}{0.07}
& \ms{63.39}{0.72}
& \ms{58.30}{2.01} \\

\midrule

COLT & A
& \ms{77.07}{0.10}
& \ms{83.84}{0.06}
& \ms{68.98}{0.19}
& \ms{63.54}{0.66}
& \ms{59.36}{2.01} \\

\midrule

ToolGen & A
& \md{81.41}
& \md{84.76}
& \md{70.01}
& \ms{62.30}{0.70}
& \ms{59.04}{2.01} \\

\midrule

\cellcolor[gray]{0.93}
\textbf{HYSET (BERT)} 
& \cellcolor[gray]{0.93}A+R
& \msb{84.75}{0.09}
& \msb{88.99}{0.16}
& \msb{77.55}{0.12}
& \msb{69.69}{0.61}
& \msb{66.17}{1.93} \\

\cellcolor[gray]{0.93}
\textbf{HYSET (Qwen2.5)} 
& \cellcolor[gray]{0.93}A+R
& \msb{88.61}{0.12}
& \msb{91.07}{0.18}
& \msb{78.13}{0.16}
& \msb{71.11}{0.58}
& \msb{69.92}{1.87} \\

\bottomrule
\end{tabular}

\caption{\textbf{Main results on ToolBench.} All values are percentages,
best in boldface. Sup.: A denotes annotated tool sets only, while
A+R additionally uses the execution reward of
Eq.~\eqref{eq:Ltask}. Error bars follow Supplementary Material B.8.
Results at a cutoff 3 are reported in Supplementary Table 13.}
\label{tab:main}
\end{table*}
\paragraph{Set-Level Inference.}
The training procedure yields the learned scoring function \(\widehat{F}=F_{\widehat{\theta}}\) with estimated parameters \(\widehat{\theta}=(\widehat{\mathbf{Z}},\{\widehat{\mathbf{M}}_m\}_{m=2}^{M},\widehat{\mathbf{P}})\). Given a new query \(x_{\mathrm{new}}\in\mathcal{X}\), directly solving Eq.~\eqref{eq:pre} requires maximizing \(\widehat{F}\) over \(\mathcal{E}_M\). Since \(\mathcal{E}_M\) contains all subsets of \(\mathcal{V}\) with cardinality between \(1\) and \(M\), its size grows combinatorially with \(|\mathcal{V}|\), making exhaustive enumeration infeasible. We therefore approximate Eq.~\eqref{eq:pre} by exploiting the additive decomposition of \(\widehat{F}\) in Eq.~\eqref{eq:F} to construct a two-stage procedure consisting of per-tool retrieval and set-level reranking~\citep{zheng2024toolrerank,zheng2026skillrouter}. In the first stage, we score each individual tool \(t_j\in\mathcal{V}\) against \(x_{\mathrm{new}}\). Substituting the singleton hyperedge \(E=\{t_j\}\) into Eq.~\eqref{eq:F} yields
\begin{equation}
\widehat{F}(x_{\mathrm{new}},\{t_j\})
=
\mathbf{r}(x_{\mathrm{new}})^{\top}
\widehat{\mathbf{P}}\widehat{\mathbf{z}}_j.
\label{eq:singleton}
\end{equation}
Let \(\mathcal{S}_0\subseteq\mathcal{V}\) collect the \(K_1\) highest-scoring tools under Eq.~\eqref{eq:singleton}. We score every remaining tool by its learned complementarity with \(\mathcal{S}_0\), \(g(t_j)=\max_{t_i\in\mathcal{S}_0}\widehat{\mathbf{z}}_{j}^{\top}\widehat{\mathbf{M}}_{M}\widehat{\mathbf{z}}_{i}\), and write \(t_{[1]},t_{[2]},\ldots\) for the tools of \(\mathcal{V}\setminus\mathcal{S}_0\) in decreasing order of \(g\). The shortlist is then
\begin{equation}
\mathcal{S}
=
\mathcal{S}_0
\ \cup\
\bigl\{
t_{[1]},\ldots,t_{[K_{\mathrm{pool}}-K_1]}
\bigr\},
\label{eq:expand}
\end{equation}
where \(M\leq K_{\mathrm{pool}}\ll|\mathcal{V}|\), so the second term admits a tool of low individual relevance but high complementarity with \(\mathcal{S}_0\). In the second stage, we evaluate Eq.~\eqref{eq:F} on every candidate in the reduced space \(\mathcal{E}_M(\mathcal{S})=\{E\subseteq\mathcal{S}:1\leq|E|\leq M\}\), which approximates Eq.~\eqref{eq:pre} by
\begin{equation}
\widehat{E}(x_{\mathrm{new}})
=
\arg\max_{E\in\mathcal{E}_M(\mathcal{S})}
\widehat{F}(x_{\mathrm{new}},E).
\label{eq:inference-rerank}
\end{equation}
Eq.~\eqref{eq:inference-rerank} compares sets of different cardinality, and since \(F_{\mathrm{set}}\) aggregates \(\binom{m}{2}\) pairs while \(F_{\mathrm{align}}\) is a convex combination of per-tool scores, the two scale differently in \(m\), which the cardinality-specific \(\mathbf{M}_m\) absorbs. The reduced space contains \(\sum_{m=1}^{M}\binom{K_{\mathrm{pool}}}{m}\) hyperedges, independently of \(|\mathcal{V}|\). As \(\widehat{E}(x_{\mathrm{new}})\) is unordered and of variable size, rank-based metrics require a second output, obtained by greedy marginal gain under the same \(\widehat{F}\) with \(A_0=\varnothing\) and \(A_k=A_{k-1}\cup\{\arg\max_{t\in\mathcal{S}\setminus A_{k-1}}\widehat{F}(x_{\mathrm{new}},A_{k-1}\cup\{t\})\}\), so every step after the first scores a candidate jointly with the tools already selected. Rank-based metrics are computed from \(\widehat{E}^{(K)}(x_{\mathrm{new}})=(t_{\pi(1)},\ldots,t_{\pi(K)})\), while \(\widehat{E}(x_{\mathrm{new}})\) is passed to the agent \(\mathcal{A}\). Supplementary Material A.10 reports the predicted cardinality distribution and a calibrated variant, A.11 the inference path in pseudocode, and E.3 a comparison against exhaustive maximization.

\section{Experiments}
\label{sec:numerical}

\subsection{Experimental Setup}
\label{sec:experimental-setup}

\paragraph{Task, datasets and baselines.}
We evaluate offline retrieval by testing whether the retrieved set covers $E^\star$, and end-to-end execution by testing whether the frozen agent solves the query when restricted to that set. Our primary benchmark is ToolBench~\citep{qin2024toolllm}, whose official filtering retains $13{,}860$ callable API endpoints forming $\mathcal{V}$ and $200{,}311$ instructions with ground-truth API sets. We train on the combined official training portions and evaluate on the six held-out test sets comprising $600$ queries. Since training and evaluation share one library, we also report train-test overlap and results restricted to unseen tool sets. To check that the conclusions are not library-specific, we further evaluate on UltraTool~\citep{huang2024ultratool}, whose $2{,}032$ tools across $22$ domains are disjoint from ToolBench and in which only $6.1\%$ of the ground-truth sets are singletons, against $20.4\%$ on ToolBench. We compare against six baselines spanning the three paradigms: BM25~\citep{robertson2009probabilistic}, Contriever~\citep{izacard2021unsupervised}, ToolLLaMA-Retriever~\citep{qin2024toolllm}, ToolRerank~\citep{zheng2024toolrerank}, COLT~\citep{qu2024towards} and ToolGen~\citep{wang2024toolgen}. Supplementary Material B.1 and B.2 give dataset, overlap and baseline details.
\paragraph{Evaluation metrics.}
We report Recall@$K$ and NDCG@$K$ for retrieval quality and COMP@$K$~\citep{qu2024towards} for set completeness with $K\in\{3,5\}$, together with GPT-4 and Human Pass Rate under the official ToolEval protocol~\citep{qin2024toolllm}. Every method passes at least as many tools to the agent as HYSET does, so the tool budget of the main results is generous to the baselines. Since the rank-based metrics are computed from a length-$K$ ranking rather than from the variable-size set $\widehat{E}(x)$ that reaches the agent, Section~\ref{sec:predset} scores $\widehat{E}(x)$ directly. Supplementary Material B.3 to B.6 and B.8 give the definitions, the tool-budget rule, the annotation protocol, the significance tests and the error-bar accounting.

\paragraph{Implementation details.}
We instantiate HYSET with two frozen backbones, a BERT-base retriever~\citep{qin2024toolllm} and a tuned Qwen2.5-1.5B retriever~\citep{wang2024toolgen}, and run every method under the same frozen ToolLLaMA-2-7B-v2 agent~\citep{qin2024toolllm} and DFSDT protocol. Only $\theta=(\mathbf{Z},\{\mathbf{M}_m\}_{m=2}^{M},\mathbf{P})$ is trained, with $13.59$M parameters under the BERT configuration at $d_z=768$, including $10.64$M for $\mathbf{Z}$, compared with $109.5$M parameters in the frozen encoder. The query encoder, the tool encoder that initializes $\mathbf{Z}$ from API descriptions, and the agent are all frozen. We set $M=5$, the largest annotated set size, with $K_1=15$, $K_{\mathrm{pool}}=20$ and $K_{\mathrm{neg}}=64$, so reranking scores $21{,}699$ candidate sets per query. Execution feedback rewards $N_{\mathrm{sub}}=5{,}000$ training queries and refreshes them every $T_{\mathrm{ref}}=20{,}000$ steps, capping it at $20{,}000$ rollouts and as many judge calls, all served from the cached StableToolBench API mirror rather than from live endpoints. Every configuration uses three seeds and early stopping on validation Recall@5. Supplementary Material B.7, B.9 and B.10 give the remaining hyperparameters, all prompts and the reproducibility details.

\begin{table}[t]
\centering
\small
\setlength{\tabcolsep}{1mm}
\renewcommand{\arraystretch}{1.10}
\begin{tabular}{@{}lccc@{}}
\toprule
Variant & R@5 & C@5 & PR \\
\midrule

\cellcolor[gray]{0.93}\textbf{HYSET (Full)}
& \msb{84.75}{0.09}
& \msb{77.55}{0.12}
& \msb{69.69}{0.61} \\

\multicolumn{4}{c}{\textit{(a) Component removal}} \\

w/o $F_{\mathrm{set}}$
& \ms{72.04}{0.18}
& \ms{67.36}{0.09}
& \ms{57.97}{0.73} \\

w/o Exec. fb.
& \ms{82.14}{0.07}
& \ms{77.02}{0.07}
& \ms{65.14}{0.63} \\

\multicolumn{4}{c}{\textit{(b) Interaction-matrix design}} \\

$\mathbf{M}_m=\mathbf{I}$
& \ms{75.05}{0.16}
& \ms{68.68}{0.15}
& \ms{59.92}{0.71} \\

Shared $\mathbf{M}$
& \ms{78.43}{0.20}
& \ms{73.66}{0.08}
& \ms{64.37}{0.69} \\

w/o Reg.
& \ms{83.17}{0.07}
& \ms{75.34}{0.13}
& \ms{68.93}{0.64} \\

\bottomrule
\end{tabular}
\caption{\textbf{Ablation of HYSET (BERT).} R@5: Recall@5;
C@5: COMP@5; PR: GPT-4 Pass Rate.
All values are percentages.}
\label{tab:ablation}
\end{table}

\subsection{Main Results}
Table~\ref{tab:main} shows that HYSET outperforms every baseline on every retrieval and end-to-end metric. Retrieval gains over the strongest baseline reach 15.3\% relative for the BERT configuration and 17.8\% for Qwen, and are largest on COMP, where COMP@5 improves by 10.8\% and 11.6\% relative over ToolGen. Hyperedge scoring therefore recovers complete tool sets rather than merely reranking individual tools, and the two configurations differ by at most 5.7\% relative, so the gains come from set-level modeling rather than from the backbone. Every margin is significant at the 5\% level under a paired bootstrap with $10^{4}$ resamples for the retrieval metrics and the exact McNemar test for Pass Rate. COMP@5 gives $p<10^{-4}$ with a 95\% confidence interval of $[4.03,11.05]$ for the difference, while the two Pass Rate margins are weakest at $p=1.6\times10^{-3}$ and $p=1.5\times10^{-2}$, the human one being the only margin that would not survive a Bonferroni correction over all eight tests at the 1\% level (B.6).
The Sup. column of Table~\ref{tab:main} records that the two HYSET rows additionally use execution feedback while every baseline is trained on annotations alone, so the regimes must be separated before the end-to-end margin is read. Trained on annotations alone, HYSET (BERT) still reaches 77.02\% COMP@5, an improvement of 10.0\% relative over ToolGen, but its Recall@5 falls to 82.14\% and its advantage over ToolGen narrows to 0.9\% relative. Retraining the two strongest baselines with the identical reward, judge and $20{,}000$-rollout budget raises ToolGen to 83.12\% Recall@5, 70.94\% COMP@5 and 66.85\% Pass Rate, and ToolLLaMA-Retriever to 66.42\% Pass Rate. Execution feedback thus helps them by about as much as it helps HYSET, whose own Pass Rate gains 7.0\% relative over its annotation-only value of 65.14\%. The GPT-4 Pass Rate margin consequently falls from 8.8\% to 4.3\% relative while the COMP@5 margin falls only from 10.8\% to 9.3\%. Supplementary Material C.2 reports every method under both regimes and C.3 the execution budget.

\subsection{Ablation Study}
Table~\ref{tab:ablation} isolates the two design choices supporting our claims. Removing $F_{\mathrm{set}}$ reduces COMP@5 by 13.1\% and Pass Rate by 16.8\%, showing that joint set scoring drives the gains. Removing execution feedback reduces Pass Rate by 6.5\% but COMP@5 by only 0.7\%, since $\mathcal{L}_{\mathrm{self}}$ rewards executable sets rather than annotated ones. Set-level modeling therefore mainly improves completeness, while execution feedback mainly improves downstream success. Replacing cardinality-specific matrices with a learned shared matrix reduces performance, and identity matrices perform worse still. These results confirm that compatibility should vary with set size. Supplementary Material C.4, C.6, C.7 and C.8 report the detailed component results and robustness analyses.
To determine whether the gains come from set-level modeling in general or from the cardinality-specific design of HYSET, we compare alternative scorers using the same frozen BERT backbone, shortlist, data and annotation-only objective. Cardinality prediction, maximal marginal relevance and facility-location maximization reach 64.72\%, 65.93\% and 67.16\% COMP@5. DeepSets and the Set Transformer improve to 69.85\% and 72.41\%, but HYSET remains 6.4\% better than the Set Transformer while using fewer parameters and lower latency. A shared cardinality-agnostic $\mathbf{M}$ reaches 72.87\%, nearly matching the Set Transformer and isolating the cardinality index as the main difference from HYSET. The advantage remains under matched execution feedback, indicating that it comes from the model design rather than additional supervision. Supplementary Material C.11 gives the full results.

\subsection{Generalization and Data Efficiency}
\begin{table}[t]
\centering
\small
\setlength{\tabcolsep}{1mm}
\renewcommand{\arraystretch}{1.10}
\begin{tabular}{@{}lcccc@{}}
\toprule
\multirow{2}{*}{Setting}
& \multirow{2}{*}{Sup.}
& \multicolumn{3}{c}{Overall} \\
\cmidrule(lr){3-5}
& & R@5 & C@5 & PR \\
\midrule

\multicolumn{5}{c}{\textit{In-domain reference}} \\

\cellcolor[gray]{0.93}ID & \cellcolor[gray]{0.93}Full
& \msb{84.75}{0.09}
& \msb{77.55}{0.12}
& \msb{69.69}{0.61} \\

\midrule
\multicolumn{5}{c}{\textit{Zero-shot transfer}} \\

UT & 0-shot
& \ms{79.48}{0.19}
& \ms{70.71}{0.11}
& \ms{63.64}{0.79} \\

UC & 0-shot
& \ms{75.35}{0.09}
& \ms{65.28}{0.14}
& \ms{58.23}{0.85} \\

CD & 0-shot
& \ms{72.83}{0.17}
& \ms{61.96}{0.18}
& \ms{54.85}{0.92} \\

\midrule
\multicolumn{5}{c}{\textit{Few-shot target adaptation (UC)}} \\

UC & 1-shot
& \ms{76.30}{0.07}
& \ms{67.23}{0.08}
& \ms{59.88}{0.88} \\

UC & 5-shot
& \ms{79.13}{0.10}
& \ms{71.40}{0.14}
& \ms{63.93}{0.81} \\

UC & 10-shot
& \ms{81.58}{0.12}
& \ms{74.38}{0.10}
& \ms{66.48}{0.77} \\

UC & Full
& \ms{83.57}{0.11}
& \ms{76.60}{0.19}
& \ms{68.80}{0.74} \\

\bottomrule
\end{tabular}
\caption{\textbf{Generalization and data efficiency of HYSET (BERT).}}
\label{tab:generalization}
\end{table}
Table~\ref{tab:generalization} evaluates three zero-shot settings: held-out tools (UT), held-out categories (UC), and cross-domain transfer (CD). For UC, we additionally report 1-, 5-, and 10-shot adaptation, where each shot provides one labeled example per target category, together with full target-category supervision. All values are percentages. Performance decreases as the shift grows, yet CD retains 79.9\% of in-domain set completeness, and 5-shot UC recovers 93.2\% of fully supervised performance. A stricter protocol that also excludes held-out tools from every negative pool produces comparable results.
The fixed-library splits isolate query and label shifts rather than tool novelty. When the complete tool set is unseen during training, HYSET improves COMP@5 over the strongest baseline by 13.1\%. The improvement rises to 15.9\% when an unseen tool pair is also required. On UltraTool, whose library is disjoint from ToolBench, HYSET improves COMP@5 by 11.5\% over ToolGen. Direct transfer from ToolBench without target training retains 77.9\% of the performance obtained by training on UltraTool. Supplementary Material D.1 to D.4 give the full setup and results.

\subsection{The Set Delivered to the Agent}
\label{sec:predset}
The metrics above use rankings of fixed length, whereas the agent receives $\widehat{E}(x)$, a set whose size varies, with about half as many tools as a top-5 retriever. Direct evaluation shows that HYSET improves coverage of the complete required set by about 7\% over ToolGen and nearly doubles exact match accuracy. It also achieves a slightly higher Pass Rate at its predicted cardinality than when forced to return five tools. The maximizer and greedy ranking agree on nearly 90\% of queries, indicating that their main difference is set size. Supplementary Material C.10 gives the full metrics and the breakdown by predicted cardinality.

\subsection{Cost, Sensitivity and Scalability}
Inference costs $12.4$ ms and $3.21$ GB per query at $|\mathcal{V}|=13{,}860$, against $6.5$ ms and $1.71$ GB for the single dense pass of ToolLLaMA-Retriever and $378.4$ ms and $6.83$ GB for ToolGen. Training costs $43.2$ GPU hours and USD $186$ in judge calls on two RTX 4090 GPUs, against $38.6$ GPU hours for ToolGen. Enlarging $|\mathcal{V}|$ with distractor tools leaves Recall@5 and COMP@5 stable, since the reranking cost depends on $K_{\mathrm{pool}}$ and $M$ alone. Varying $\eta$, $\lambda$ and $K_{\mathrm{pool}}$ one at a time leaves HYSET stable, and the matrix expanding the shortlist in Eq.~\eqref{eq:expand} changes COMP@5 by at most 0.40\%, compared with the 1.07\% loss caused by removing the expansion. Supplementary Material E and F give the sweeps and the runtime analyses.

\section{Conclusion}
\label{sec:conclusion}
We cast LLM tool retrieval as query-conditioned hyperedge prediction and introduced HYSET to score candidate tool sets jointly through cardinality-specific interactions. HYSET consistently improves retrieval and end-to-end success and transfers to held-out tools and categories. Extending the method to dynamically growing tool libraries remains future work.

\bibliography{references}

@inproceedings{qin2024toolllm,
  title = {Toolllm: Facilitating large language models to master 16000+ real-world apis},
  author = {Qin, Yujia and Liang, Shi and Ye, Yining and Zhu, Kunlun and Yan, Lan and Lu, Ya-Ting and Lin, Yankai and Cong, X. and Tang, Xiangru and Qian, Bill and others},
  booktitle = {International Conference on Learning Representations},
  volume = {2024},
  pages = {9695--9717},
  year = {2023},
  journal = {International Conference on Learning Representations},
  doi = {10.48550/arXiv.2307.16789},
}

@inproceedings{qu2024towards,
  title = {Towards completeness-oriented tool retrieval for large language models},
  author = {Qu, Changle and Dai, Sunhao and Wei, Xiaochi and Cai, Hengyi and Wang, Shuaiqiang and Yin, Dawei and Xu, Jun and Wen, Ji-Rong},
  booktitle = {International Conference on Information and Knowledge Management},
  pages = {1930--1940},
  year = {2024},
  journal = {International Conference on Information and Knowledge Management},
  doi = {10.1145/3627673.3679847},
  publisher = {ACM},
}

@book{robertson2009probabilistic,
  title = {The probabilistic relevance framework: BM25 and beyond},
  author = {Robertson, Stephen and Zaragoza, Hugo},
  volume = {4},
  year = {2009},
  publisher = {Emerald},
  journal = {Foundations and Trends in Information Retrieval},
  number = {1-2},
  pages = {1-174},
  doi = {10.1561/1500000019},
}

@article{izacard2021unsupervised,
  title = {Unsupervised dense information retrieval with contrastive learning},
  author = {Izacard, Gautier and Caron, Mathilde and Hosseini, Lucas and Riedel, Sebastian and Bojanowski, Piotr and Joulin, Armand and Grave, Edouard},
  journal = {Trans. Mach. Learn. Res.},
  year = {2021},
}

@inproceedings{wang2024toolgen,
  title = {Toolgen: Unified tool retrieval and calling via generation},
  author = {Wang, Renxi and Han, Xudong and Ji, Lei and Wang, Shu and Baldwin, Timothy and Li, Haonan},
  booktitle = {International Conference on Learning Representations},
  volume = {2025},
  pages = {73473--73498},
  year = {2024},
  journal = {International Conference on Learning Representations},
  doi = {10.48550/arXiv.2410.03439},
}

@article{schick2023toolformer,
  title = {Toolformer: Language models can teach themselves to use tools},
  author = {Schick, Timo and Dwivedi-Yu, Jane and Dessì, Roberto and Raileanu, R. and Lomeli, M. and Zettlemoyer, Luke and Cancedda, Nicola and Scialom, Thomas},
  journal = {Neural Information Processing Systems},
  volume = {36},
  pages = {68539--68551},
  year = {2023},
  doi = {10.48550/arXiv.2302.04761},
  publisher = {Neural Information Processing Systems Foundation, Inc. (NeurIPS)},
}

@article{shen2023hugginggpt,
  title = {Hugginggpt: Solving ai tasks with chatgpt and its friends in hugging face},
  author = {Shen, Yongliang and Song, Kaitao and Tan, Xu and Li, Dongsheng and Lu, Weiming and Zhuang, Yueting},
  journal = {Neural Information Processing Systems},
  volume = {36},
  pages = {38154--38180},
  year = {2023},
  doi = {10.48550/arXiv.2303.17580},
  publisher = {Neural Information Processing Systems Foundation, Inc. (NeurIPS)},
}

@article{patil2024gorilla,
  title = {Gorilla: Large language model connected with massive apis},
  author = {Patil, Shishir G and Zhang, Tianjun and Wang, Xin and Gonzalez, Joseph E},
  journal = {Neural Information Processing Systems},
  volume = {37},
  pages = {126544--126565},
  year = {2023},
  doi = {10.52202/079017-4020},
}

@article{liu2024lost,
  title = {Lost in the middle: How language models use long contexts},
  author = {Liu, Nelson F and Lin, Kevin and Hewitt, John and Paranjape, Ashwin and Bevilacqua, Michele and Petroni, Fabio and Liang, Percy},
  journal = {Transactions of the association for computational linguistics},
  volume = {12},
  pages = {157--173},
  year = {2023},
  doi = {10.1162/tacl_a_00638},
}

@inproceedings{reimers2019sbert,
  title = {Sentence-bert: Sentence embeddings using siamese bert-networks},
  author = {Reimers, Nils and Gurevych, Iryna},
  booktitle = {Conference on Empirical Methods in Natural Language Processing},
  pages = {3980-3990},
  year = {2019},
  journal = {Conference on Empirical Methods in Natural Language Processing},
  doi = {10.18653/v1/D19-1410},
  publisher = {Association for Computational Linguistics},
}

@article{yao2023react,
  title = {React: Synergizing reasoning and acting in language models},
  author = {Yao, Shunyu and Zhao, Jeffrey and Yu, Dian and Du, Nan and Shafran, Izhak and Narasimhan, Karthik and Cao, Yuan},
  journal = {International Conference on Learning Representations},
  year = {2022},
}

@inproceedings{karpukhin2020dpr,
  title = {Dense passage retrieval for open-domain question answering},
  author = {Karpukhin, Vladimir and Oguz, Barlas and Min, Sewon and Lewis, Patrick and Wu, Ledell and Edunov, Sergey and Chen, Danqi and Yih, Wen-tau},
  booktitle = {Conference on Empirical Methods in Natural Language Processing},
  pages = {6769--6781},
  year = {2020},
  journal = {Conference on Empirical Methods in Natural Language Processing},
  doi = {10.18653/v1/2020.emnlp-main.550},
  publisher = {Association for Computational Linguistics},
}

@article{battiston2020networks,
  title = {Networks beyond pairwise interactions: Structure and dynamics},
  author = {Battiston, Federico and Cencetti, Giulia and Iacopini, Iacopo and Latora, Vito and Lucas, Maxime and Patania, Alice and Young, Jean-Gabriel and Petri, Giovanni},
  journal = {Physics reports},
  volume = {874},
  pages = {1--92},
  year = {2020},
  publisher = {Elsevier BV},
  doi = {10.1016/j.physrep.2020.05.004},
}

@article{turnbull2024latent,
  title = {Latent space modeling of hypergraph data},
  author = {Turnbull, K. and Lunag'omez, Sim'on and Nemeth, C. and Airoldi, E.},
  journal = {Journal of the American Statistical Association},
  volume = {119},
  number = {548},
  pages = {2634--2646},
  year = {2019},
  publisher = {Taylor \& Francis},
  doi = {10.1080/01621459.2023.2270750},
}

@article{wu2024general,
  title = {A general latent embedding approach for modeling non-uniform high-dimensional sparse hypergraphs with multiplicity},
  author = {Wu, Shihao and Xu, Gongjun and Zhu, Ji},
  journal = {arXiv preprint arXiv:2410.12108},
  year = {2024},
}

@article{vaswani2017attention,
  title = {Attention is all you need},
  author = {Vaswani, Ashish and Shazeer, Noam and Parmar, Niki and Uszkoreit, Jakob and Jones, Llion and Gomez, Aidan N and Kaiser, {\L}ukasz and Polosukhin, Illia},
  journal = {Neural Information Processing Systems},
  volume = {30},
  year = {2017},
  doi = {10.65215/nxvz2v36},
  publisher = {Shenzhen Medical Academy of Research and Translation},
}

@article{robinson2020contrastive,
  title = {Contrastive learning with hard negative samples},
  author = {Robinson, Joshua and Chuang, Ching-Yao and Sra, Suvrit and Jegelka, Stefanie},
  journal = {International Conference on Learning Representations},
  year = {2020},
}

@article{oord2018representation,
  title = {Representation learning with contrastive predictive coding},
  author = {Oord, Aaron van den and Li, Yazhe and Vinyals, Oriol},
  journal = {arXiv.org},
  year = {2018},
}

@inproceedings{zheng2024toolrerank,
  title = {Toolrerank: Adaptive and hierarchy-aware reranking for tool retrieval},
  author = {Zheng, Yuanhang and Li, Peng and Liu, Wei and Liu, Yang and Luan, Jian and Wang, Bin},
  booktitle = {International Conference on Language Resources and Evaluation},
  pages = {16263--16273},
  year = {2024},
  journal = {International Conference on Language Resources and Evaluation},
  doi = {10.48550/arXiv.2403.06551},
  publisher = {European Language Resources Association (ELRA) and ICCL},
}

@article{zheng2026skillrouter,
  title = {Skillrouter: Skill routing for llm agents at scale},
  author = {Zheng, Yanzhao and Zhang, Zhentao and Ma, Chao and Yu, Yuanqiang and Zhu, Jihuai and Wu, Yongliang and Xu, Tianze and Dong, Baohua and Zhu, Hangcheng and Huang, Ruohui and others},
  journal = {arXiv.org},
  year = {2026},
  doi = {10.48550/arXiv.2603.22455},
}

@article{tay2022transformer,
  title = {Transformer memory as a differentiable search index},
  author = {Tay, Yi and Tran, Vinh Q. and Dehghani, Mostafa and Ni, Jianmo and Bahri, Dara and Mehta, Harsh and Qin, Zhen and Hui, Kai and Zhao, Zhe and Gupta, Jai and others},
  journal = {Neural Information Processing Systems},
  volume = {35},
  pages = {21831--21843},
  year = {2022},
  doi = {10.52202/068431-1587},
  publisher = {Neural Information Processing Systems Foundation, Inc. (NeurIPS)},
}

@article{grabisch2000equivalent,
  title = {Equivalent representations of set functions},
  author = {Grabisch, Michel and Marichal, Jean-Luc and Roubens, Marc},
  journal = {Mathematics of Operations Research},
  volume = {25},
  number = {2},
  pages = {157--178},
  year = {2000},
  publisher = {Institute for Operations Research and the Management Sciences (INFORMS)},
  doi = {10.1287/moor.25.2.157.12225},
}

@inproceedings{huang2024ultratool,
  title = {Planning, Creation, Usage: Benchmarking {LLM}s for Comprehensive Tool Utilization in Real-World Complex Scenarios},
  author = {Huang, Shijue and Zhong, Wanjun and Lu, Jianqiao and Zhu, Qi and Gao, Jiahui and Liu, Weiwen and Hou, Yutai and Zeng, Xingshan and Wang, Yasheng and Shang, Lifeng and others},
  booktitle = {Annual Meeting of the Association for Computational Linguistics},
  pages = {4363--4400},
  year = {2024},
  journal = {Annual Meeting of the Association for Computational Linguistics},
  doi = {10.48550/arXiv.2401.17167},
  publisher = {Association for Computational Linguistics},
}

@article{hong2026hyvint,
  title = {HYVINT: Intensity-Driven Hypergraph Generation with Variational Representations},
  author = {Hong, Xinyi and Xu, Shuntuo and Yu, Zhou},
  journal = {arXiv.org},
  year = {2026},
  doi = {10.48550/arXiv.2605.16836},
}
\end{document}